\documentclass[conference,letterpaper]{IEEEtran}

\IEEEoverridecommandlockouts

\newcommand{\CLASSINPUTtoptextmargin}{19mm}
\newcommand{\CLASSINPUTbottomtextmargin}{18mm}
\newcommand{\CLASSINPUTinnersidemargin}{19mm}
\newcommand{\CLASSINPUToutersidemargin}{19mm}

\usepackage[export]{adjustbox}
\usepackage{etoolbox}

\usepackage{times}
\usepackage{epsfig}
\usepackage{graphicx}
\usepackage{amsmath}
\usepackage{amssymb}

\usepackage{dblfloatfix}
\usepackage{booktabs, tabularx, siunitx,pifont}
\usepackage{enumitem}
\usepackage{algpseudocode}
\usepackage{algorithm}

\usepackage{xspace}
\makeatletter
\DeclareRobustCommand\onedot{\futurelet\@let@token\@onedot}
\def\@onedot{\ifx\@let@token.\else.\null\fi\xspace}
\def\eg{\emph{e.g}\onedot} 
\def\ie{\emph{i.e}\onedot} 
 
 \def\vs{\emph{vs}\onedot}
 
\def\etal{\emph{et al}\onedot}
\makeatother

\algrenewcommand\algorithmicrequire{\textbf{Input}}
\algrenewcommand\algorithmicensure{\textbf{Output}}

\usepackage[pagebackref=true,breaklinks=true,letterpaper=true,colorlinks,bookmarks=false]{hyperref}

\makeatletter
\patchcmd{\@makecaption}
  {\scshape}
  {}
  {}
  {}
\makeatletter
\patchcmd{\@makecaption}
  {\\}
  {.\ }
  {}
  {}
\makeatother
\def\tablename{Table}

\makeatletter
\newcommand{\hydash}{\penalty\@M-\hskip\z@skip}
\makeatother
\title{\LARGE \bf
\vspace{6mm} L2E: Lasers to Events for 6-DoF Extrinsic Calibration of Lidars and Event Cameras}

\author{Kevin Ta$^1$ \quad
    David Bruggemann$^1$ \quad
    Tim Brödermann$^1$ \quad
    Christos Sakaridis$^1$ \quad
    Luc Van Gool$^{1,2}$
\thanks{This work was supported by the ETH Future Computing Laboratory (EFCL), financed by a donation from Huawei Technologies.}%
\thanks{$^1$Kevin Ta, David Bruggemann, Tim Brödermann, Christos Sakaridis, and Luc Van Gool are with the Computer Vision Lab, ETH Zurich, 8092 Zurich, Switzerland. {\tt\footnotesize \href{mailto:contact@kevinta.dev}{contact@kevinta.dev}}}%
\thanks{$^2$Luc Van Gool is with the Department of Electrical Engineering, KU Leuven, 3000 Leuven, Belgium.}%
}

\begin{document}

\maketitle
\thispagestyle{empty}
\pagestyle{empty}

\begin{abstract}
As neuromorphic technology is maturing, its application to robotics and autonomous vehicle systems has become an area of active research.
In particular, event cameras have emerged as a compelling alternative to frame-based cameras in low-power and latency-demanding applications.
To enable event cameras to operate alongside staple sensors like lidar in perception tasks, we propose a direct, temporally-decoupled extrinsic calibration method between event cameras and lidars. The high dynamic range, high temporal resolution, and low-latency operation of event cameras are exploited to directly register lidar laser returns, allowing information-based correlation methods to optimize for the 6-DoF extrinsic calibration between the two sensors. This paper presents the first direct calibration method between event cameras and lidars, removing dependencies on frame-based camera intermediaries and/or highly-accurate hand measurements. 
Code: \url{https://github.com/kev-in-ta/l2e}
\end{abstract}

\section{Introduction}

In the last decade, the development of many robotics applications\textemdash \eg autonomous vehicles\textemdash has accelerated due to both academic and industrial research. 
The choice of sensors is pivotal for optimizing the perception of a robot's surroundings.
Besides frame-based camera and radar, lidar has become a core perception technology for autonomous robots, drones, and vehicles.
By detecting the reflection of emitted laser pulses, lidars generate 3D point cloud representations of a scene.

More recently, event cameras have been explored as a potential alternative to traditional frame-based cameras, offering advantages in low-power and latency-demanding applications.
Event-based vision derives from neuromorphic engineering, aimed at replicating fundamental, biological neural functions~\cite{Gallego2022}. An event camera asynchronously extracts individual pixel-wise events that correspond to luminosity changes. This contrasts with traditional cameras that capture entire frames at regular intervals, even in fully static scenes. 

\begin{figure}[t]
  \centering
  \setlength{\tabcolsep}{2pt}
  \begin{tabularx}{\textwidth}{lr}
    \includegraphics[ width=0.49\columnwidth]{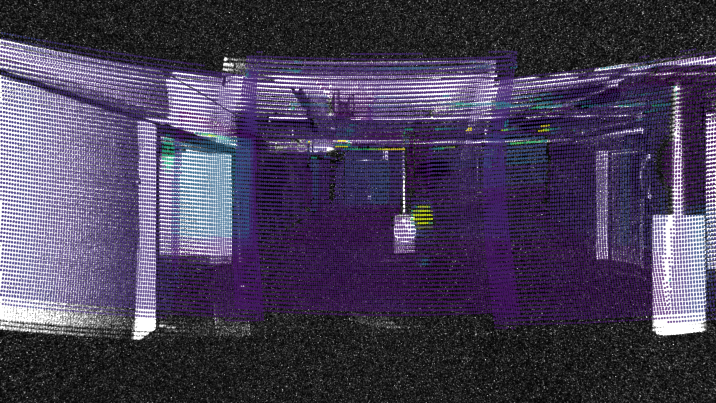} &
    \includegraphics[ width=0.49\columnwidth]{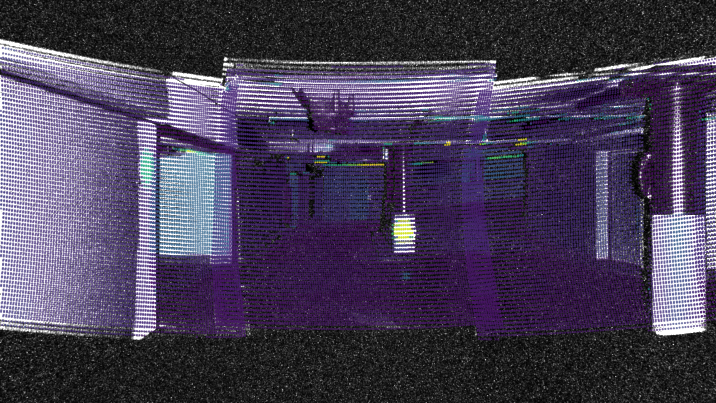} \\
    \includegraphics[ width=0.49\columnwidth]{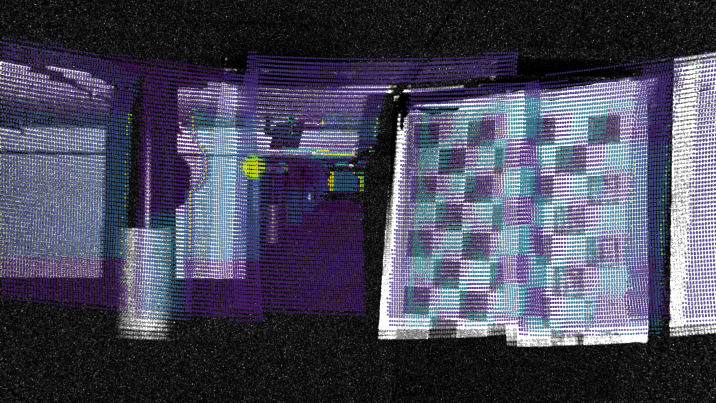} &
    \includegraphics[ width=0.49\columnwidth]{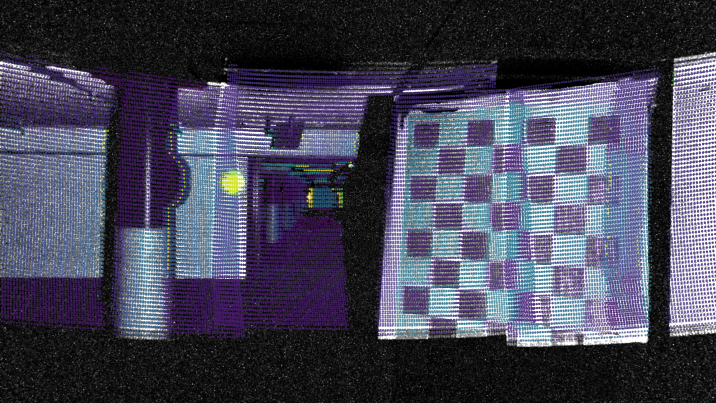} \\
    \includegraphics[ width=0.49\columnwidth]{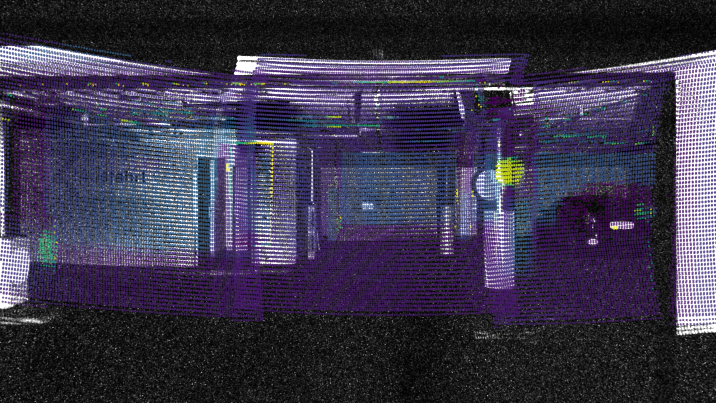} &
    \includegraphics[ width=0.49\columnwidth]{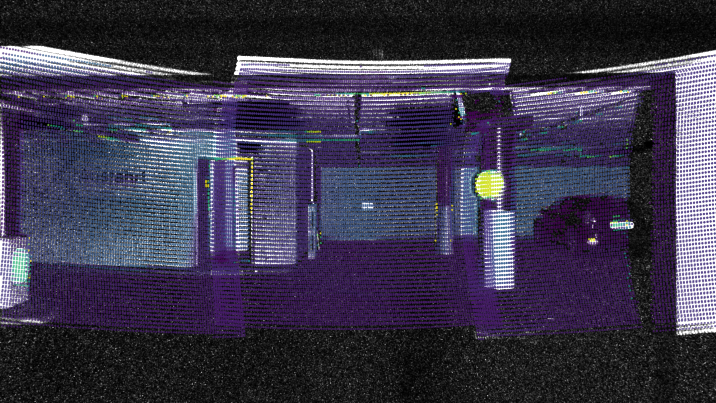}
  \end{tabularx}
  \caption{(Left) Uncalibrated scene showing improperly projected lidar points. (Right) Calibrated scene using L2E showing improved lidar projection alignment. Accumulated event maps are shown in grayscale and lidar points are overlaid in colour.} \label{fig:calibrated}
\end{figure}

To maximize performance and robustness, fundamental robot tasks such as perception, localization, and odometry often rely on multiple sensors, whose outputs are fused~\cite{Jusoh2020AApplications}.
Importantly, multiple sensors offer complementary information, \eg adverse weather conditions corrupt event camera and lidar signals to different degrees.
This said, a key aspect of enabling these sensors is to accurately calibrate their intrinsic and extrinsic parameters. These parameters specify how each modality represents the environment and how each sensor is positioned relative to the others~\cite{Yeong2021}. Without calibration, sensor fusion methods incorrectly associate spatial features, as seen in Fig.~\ref{fig:calibrated}, which negatively impacts downstream perception. 

As event cameras represent a still immature technology, methods have yet to be fully explored for the direct extrinsic calibration between them and other perception sensors. 
Current event camera calibration methods involve confounding factors such as time synchronization for temporal correction, dynamic target handling introducing elastic deformations~\cite{Hagemann2022ModelingCalibration} and motion artifacts, dynamic scenes requiring ego-motion compensation, or calibration intermediaries introducing propagation errors.
Instead, we argue that reliable and repeatable alignment of physical sensors benefits from \emph{strong environmental controls}\textemdash such as static indoor scenes and temporal decoupling.
Accordingly, we explore extrinsic lidar-event camera calibration in a fully static environment, thereby eliminating potential confounding factors.

Despite the increasing interest in event cameras, few large-scale multi-sensor datasets include event cameras to date.
To the best of our knowledge, there exist only two datasets\textemdash both automotive\textemdash which contain both lidar and event cameras: MVSEC~\cite{Zhu2018ThePerception} and DSEC~\cite{Gehrig2021}. However, neither performs direct calibration between event cameras and lidars.


In this paper we present L2E, a novel automatic pipeline using information-based optimization for calibrating a lidar with respect to an event-based camera. It leverages event-based structured light for static correspondence matching. In particular, we show how a dense lidar scan collected by a RoboSense RS-LiDAR-M1 can be easily and reliably correlated to a series of events registered by a Prophesee GEN4.1 event camera, and how we can achieve robust calibration using accumulated event maps. 
Our key contributions can be summarized as follows:
\begin{enumerate}[topsep=6pt,itemsep=3pt,partopsep=3pt, parsep=3pt]
    \item We present the first direct and temporally-decoupled 6-DoF calibration between an event camera and a lidar using accumulated event maps and lidar point clouds. 
    \item We propose a novel approach for synthesizing a static scene representation through accumulated event maps from the lidar-correlated event activity.
    \item We investigate how scene selection and scene count affect calibration accuracy, repeatability, and computational performance.
\end{enumerate}

\section{Related Work}

Neuromorphic sensors have rarely been integrated in the perception stack of autonomous robots. They suffer from a lack of maturity in neuromorphic technology. Yet, event cameras offer distinct advantages beyond the current suite of standard sensors. 
With their high temporal resolution and dynamic range, these sensors can operate in high-speed and over-saturated environments that may pose problems for frame-based cameras~\cite{Gallego2022}. The event-based paradigm has the potential to reduce data bandwidth in static scenes without sacrificing the advantageous dense representation of imaging technology. 

Recently, new datasets have emerged that incorporate event camera systems. There are two datasets containing event cameras and lidar sensors\textemdash both automotive. MVSEC~\cite{Zhu2018ThePerception} uses two DAVIS346 event cameras and a Velodyne VLP-16 lidar. The DAVIS346 camera captures a relatively low spatial resolution (346 \texttimes\ 260). DSEC~\cite{Gehrig2021} uses two Prophesee GEN3.1 event cameras and a Velodyne VLP-16 lidar. The Prophesee GEN3.1 operates at a higher spatial resolution (640 \texttimes\ 480) compared to the DAVIS346 camera, but at a lower spatial resolution than the current GEN4.1 sensors (1280 \texttimes\ 720). 
 
For calibration, MVSEC attempted to use the grayscale image produced by the DAVIS346's Active Pixel Sensor (APS) with the Camera and Range Calibration Toolbox~\cite{Geiger2012AutomaticShot}. The APS design is able to simultaneously act as a monochrome frame-based and event camera. However, the calibration results were found to be inaccurate, and the authors resorted to CAD measurements and manual adjustment for extrinsic parameters. DSEC uses the Prophesee event cameras which do not have the APS. Instead, rotation-only calibration is achieved using stereo methods and indirect calibration. Both datasets take fixed translation parameters from CAD models and solely calibrate for the 3-DoF rotation.

Intrinsic and stereo calibration of cameras is a well-established field with widely available tools such as OpenCV~\cite{opencv_library} or Kalibr~\cite{Oth2013RollingCalibration}. 
Extensions of these methods have recently been made for event cameras where state-of-the-art event-to-video reconstruction is used to interpolate frames of moving checkerboard patterns for the purpose of intrinsic and stereo calibration~\cite{Muglikar2021}. DSEC uses this image reconstruction method to employ stereo calibration between the event and frame-based cameras using the Kalibr toolbox. The authors then perform extrinsic calibration between a frame-based camera stereo pair and a lidar sensor by performing modified point-to-plane ICP~\cite{Chen1991ObjectImages} from the stereo point cloud generated by SGM~\cite{Hirschmuller2008StereoInformation} and the lidar point cloud. 

Calibration methods have been explored quite intensively between traditional cameras and lidars. These include automatic mutual information maximization schemes~\cite{Pandey2014, Taylor2012}, edge alignment methods~\cite{Zhou2018AutomaticCorrespondences, Kang2020AutomaticModel}, hand-selected point correspondences~\cite{Scaramuzza2007ExtrinsicScenes,Unnikrishnan-2005-9235}, and plane alignment methods~\cite{Geiger2012AutomaticShot, Zhou2018AutomaticCorrespondences}. These methods vary from being completely unstructured, semi-structured~\cite{An2020}, to using manufactured targets, such as standard planar checkerboards~\cite{Zhang2000ACalibration} or custom multi-modal targets~\cite{Domhof2021}. 
These methods are designed with traditional camera images in mind and are not directly applicable to an event-based data stream. 

Our basic observation is that lidar laser returns can trigger events, generating highly correlated signals in both sensors. Previous work has explored the use of event-based structured light in the visible spectrum using line or raster-pattern scans for 3D reconstruction~\cite{Brandli2014AdaptiveSensor, Muglikar2021ESL:Light, Matsuda2015MC3D:Scanning} and stereo calibration~\cite{Martel2018AnSensors}. We can naturally formulate the one-to-one correspondence matching under a mutual information framework, such as the one proposed by Pandey \etal~\cite{Pandey2014}.
Our work thus extends the use of mutual information frameworks from traditional cameras to event cameras.

\section{Sensors Overview}

\subsection{Characteristics}

\begin{figure}
  \centering
  \includegraphics[width=\columnwidth]{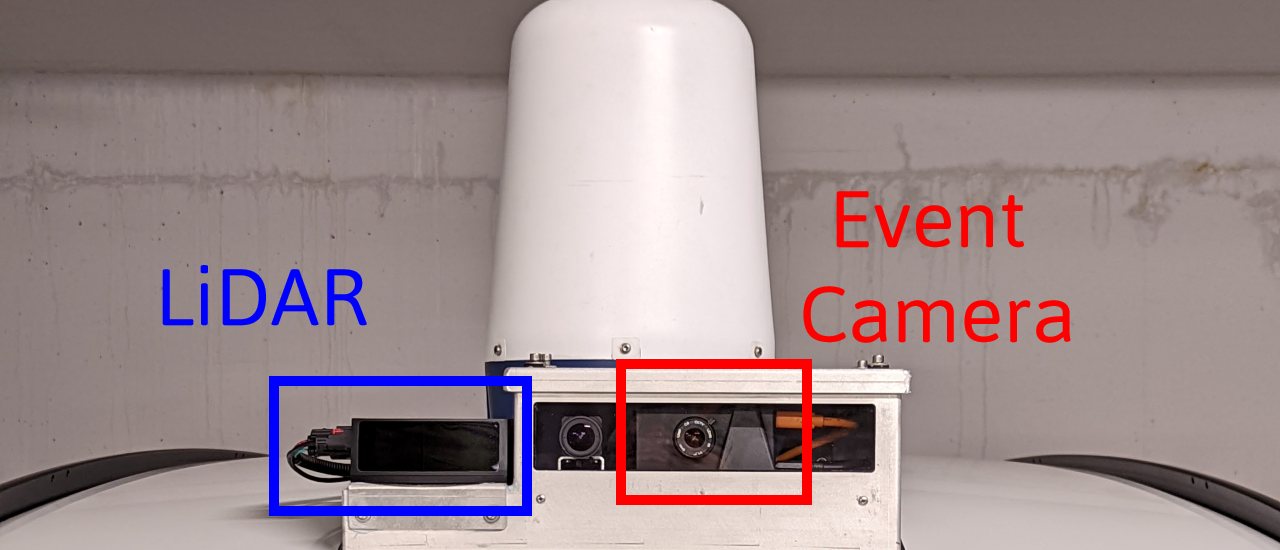}
  \caption{Annotated image of the as-built sensor platform.} \label{fig:mount}
\end{figure}

Our experimental sensor setup consists of a RoboSense RS-LiDAR-M1 MEMS lidar~\cite{RoboSense2021RS-LiDAR-M1EN} and a Prophesee GEN4.1 event camera ~\cite{Prophesee2021PropheseeBrochure}. These sensors are mounted on top of a car, facing forward. Their nominal parameters are listed in Table~\ref{tab:sensors} and the sensor installation is shown in Fig.~\ref{fig:mount}.

\begin{table}
\caption{Specifications of the MEMS lidar and event camera.} \label{tab:sensors}
\centering
\renewcommand{\arraystretch}{1.3}
\begin{tabular*}{\columnwidth}{@{\extracolsep{\fill}}lrrrr@{}}
\toprule
Sensor & Resolution & HFoV & VFoV & Freq. \\
\midrule
RS-LiDAR-M1  & 75k points & 120$^\circ$ & 25$^\circ$ & 10\,Hz \\
GEN4.1  & 1280 \texttimes\ 720 & 63$^\circ$ & 38$^\circ$ & 1.06\,Gev/s \\
\bottomrule
\end{tabular*}
\end{table}

The Prophesee GEN4.1 event-based vision system utilizes a state-of-the-art event vision sensor~\cite{Finateu20201280720Pipeline} and is capable of higher spatial resolution and smaller pixel sizes (4.86 \texttimes\ 4.86\,\si{\micro\m}) than the previous generation (15 \texttimes\ 15\,\si{\micro\m})~\cite{Prophesee2019PropheseeBrief}.
This modern event-driven sensor can operate between 5 and 100,000\,\si{lux}, with a full dynamic range in excess of 110\,dB.

Such silicon-based image sensors are naturally sensitive to wavelengths up to 1000\,\si{nm}~\cite{Gouveia2016AdvancesSensors}. Modern event cameras also utilize IR-correcting lenses as opposed to IR cut filters that would otherwise inhibit the sensor quantum efficiency (QE) and dynamic range. Additional improvement such as back-illumination also improve NIR sensitivity~\cite{Fereyre2012BackVision, Taverni2018FrontComparison}. As the most common industrial lidar wavelength is 905\,\si{nm} due to its robustness in adverse conditions~\cite{Wojtanowski2014ComparisonConditions}, most commercial lidars are capable of direct registration by modern event cameras.

The RS-LiDAR-M1 is a 905\,\si{nm} MEMS lidar that captures a large frontal field-of-view (120° horizontal) while maintaining a very small profile. With an angular resolution of 0.2°, the generated point cloud densely overlaps the event camera view.
The Prophesee GEN4.1 event camera can thus directly register the laser signals generated by the RS-LiDAR-M1. 

\subsection{Intrinsic Calibration}

The RoboSense lidar is calibrated by the manufacturer and assumed to be accurate.
For the event camera, we employ the intrinsic calibration process described in~\cite{Muglikar2021}. Accordingly, a moving target is captured as a series of events and then reconstructed as video frames with E2VID~\cite{Rebecq2019Events-to-video:Cameras, Rebecq2021HighCamera}.
The frames enable standard camera calibration using OpenCV~\cite{opencv_library}.
The employed target is a commercially-available metrology-grade checkerboard, with a reported accuracy within 0.1\,\si{mm} + 0.3\,\si{mm/m} (at 20°C). 
We calibrate using the standard pinhole camera model and the Brown–Conrady distortion model for calibration. The pinhole model, $\mathbf{K}$, is described by the focal length $(F_x,F_y)$ and principal point $(C_x,C_y)$ parameters.


\section{Event Camera-Lidar Extrinsic Calibration}

In this section, our proposed event camera-lidar extrinsic calibration method\textemdash L2E\textemdash is described in detail.
To represent extrinsic calibrations as homogeneous transformations, we collapse the transformations down to a 6-parameter vector given by the translation parameters $(x,y,z)$ and the rotational vector (axis-angle) representation $(v_1,v_2,v_3)$ where the $\text{norm}\:||\,\boldsymbol{v}\,||$ is the angle of rotation in radians. Thus the full representation of the extrinsic transformation, $\Theta$, is given by $(x, y, z, v_1, v_2, v_3)$. The translation vector will be referred to as $\boldsymbol{t}$. The expanded rotation matrix will be referred to as $\mathbf{R}$.

\subsection{Accumulated Event Map}

The events registered by the event camera are provided in the following format: $(t_e,x,y,\pm)$. Events are recorded with the time of the event, the location of the event, and the polarity of the event\textemdash\ie  whether the intensity has increased or decreased. Passing over an edge triggers a single positive (dark-to-bright) or negative (bright-to-dark) polarity event. In contrast, a laser pulse triggers both a positive and a negative event. As a lidar continuously scans a scene, we can simply increment the location for every triggered event, regardless of polarity, to get a corresponding accumulated event map, $\mathcal{E}$.

\begin{figure}
  \centering
  \includegraphics[width=\columnwidth]{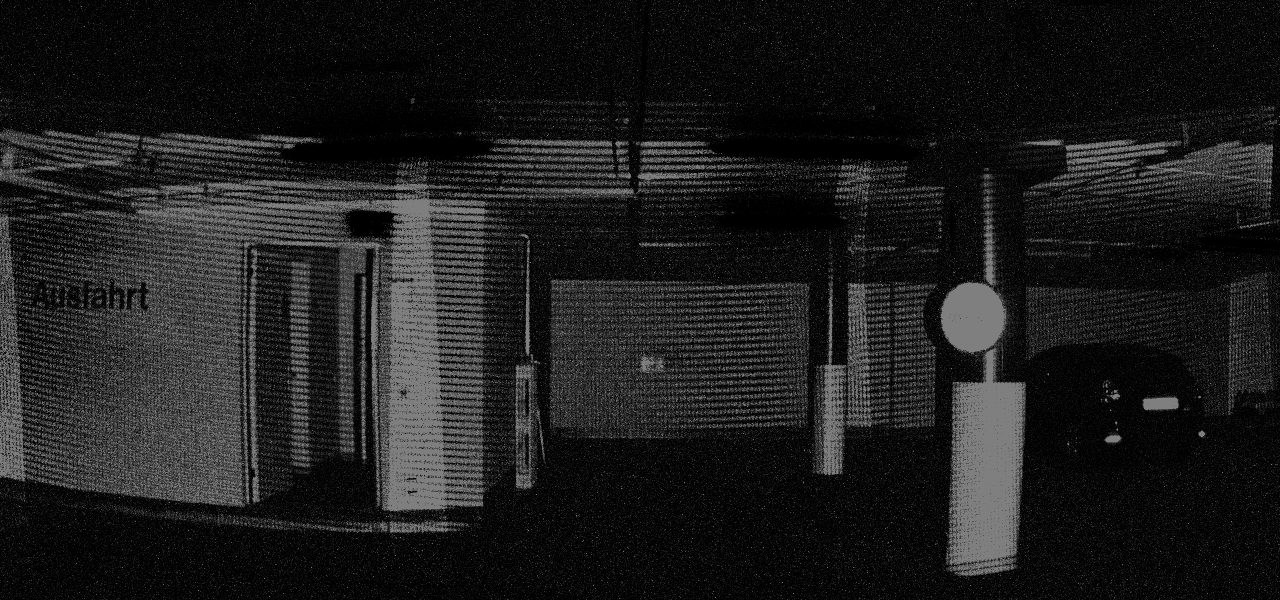}
  \includegraphics[width=\columnwidth]{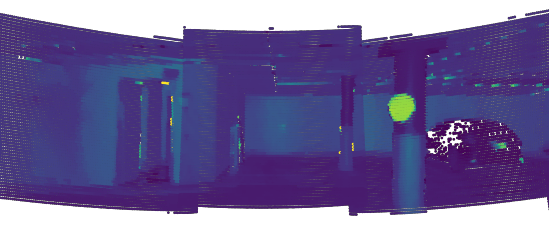}
  \caption{(Top) Accumulated event map gathered over three seconds. (Bottom) Cropped projected lidar scans. The accumulated events in the static scene are correlated with the laser returns from lidar.} \label{fig:78}
\end{figure}

After accumulating events for 3 seconds, we clip the event map such that $\mathcal{E}(x,y) \in [0,127]$ to prevent scaling issues with abnormally high event activity at a pixel location. Additionally, we apply Gaussian smoothing to the accumulated event map for smoother optimization. An accumulated event map and its corresponding lidar projection are shown in Fig.~\ref{fig:78}.

\subsection{Mutual Information}

As events are directly triggered by the lidar returns, the correlation between the lidar point cloud and accumulated event map can be exploited with mutual information optimization.

Mutual information (MI) is a measure of statistical dependence between random variables, indicating how much information one variable contains regarding the other. MI can be described in multiple ways, but we take the same entropy-based representation used in~\cite{Pandey2014}. MI is defined in terms of the entropy of the random variables $\mathrm{L}$ and $\mathrm{E}$, and their respective joint entropy $\mathrm{H}(\mathrm{L},\mathrm{E})$:
\begin{equation}
\label{eqn:mi}
    \mathrm{MI}(\mathrm{L}, \mathrm{E}) = \mathrm{H}(\mathrm{L}) + \mathrm{H}(\mathrm{E}) - \mathrm{H}(\mathrm{L},\mathrm{E})  
\end{equation}
The entropy denotes a measure of uncertainty within one variable, while the joint entropy represents the uncertainty present in the event of a co-observation of $\mathrm{L}$ and $\mathrm{E}$. We take the random variables $\mathrm{L}$ and $\mathrm{E}$ to be the lidar return intensities and the event activity in the accumulated event map, respectively. 
The entropies of random variables $\mathrm{L}$ and $\mathrm{E}$ and their joint entropy are described in Eq.~(\ref{eqn:hx}, \ref{eqn:hy}, \ref{eqn:hxy}) from their respective probabilities $\mathrm{p}_\mathrm{L}$, $\mathrm{p}_\mathrm{E}$, and $\mathrm{p}_\mathrm{LE}$. 

\begin{equation}
\label{eqn:hx}
    \mathrm{H}(\mathrm{L}) = -\sum_{l\in \mathrm{L}}\mathrm{p}_\mathrm{L}(l)\log\,\mathrm{p}_\mathrm{L}(l)
\end{equation}

\begin{equation}
\label{eqn:hy}
    \mathrm{H}(\mathrm{E}) = -\sum_{e\in \mathrm{E}}\mathrm{p}_\mathrm{E}(e)\log\,\mathrm{p}_\mathrm{E}(e)
\end{equation}

\begin{equation}
\label{eqn:hxy}
    \mathrm{H}(\mathrm{L},\mathrm{E}) = -\sum_{l\in \mathrm{L}}\sum_{e \in \mathrm{E}}\mathrm{p}_\mathrm{LE}(l,e)\log\,\mathrm{p}_\mathrm{LE}(l,e)
\end{equation}

\subsection{Probability Distribution Approximation}

We approximate the probability distribution from the intensity and activity histograms of the lidar scan and accumulated event map respectively, as was done in~\cite{Pandey2014}. Let $\mathcal{P} = \{\mathbf{P}_i: i=1,2,...,n\}$ be the set of 3D points in the 3D scan and ${\{\mathrm{L}_i: i=1,2,...,n\}}$ be the set of intensity returns for each point in the set $\mathcal{P}$. The lidar points can be projected into the image space through the searched-for extrinsics ($\mathbf{R}$, $\boldsymbol{t}$) and the known intrinsics $\mathbf{K}$. 
\begin{equation}
\label{eqn:proj}
    \boldsymbol{p}_i = \mathbf{K}[\mathbf{R}|\boldsymbol{t}]\mathbf{P}_i
\end{equation}
The location of the projected point, $\boldsymbol{p}_i$, is then used to retrieve the associated event activity in the accumulated event map image $\mathcal{E}$ as shown in Eq.~(\ref{eqn:grayscale}).
\begin{equation}
\label{eqn:grayscale}
    \mathrm{E}_i = \mathcal{E}(\boldsymbol{p}_i)
\end{equation}
Through accumulating 2D projected points that lie within the camera view, histograms $\mathrm{hist}_\mathrm{L}$ and $\mathrm{hist}_\mathrm{E}$ are generated from the discretized lidar intensities of each point and the event activity at the projected location. A joint 2D histogram is also generated from each intensity-activity pair. This process is detailed in Algorithm \ref{alg:hist}.

\begin{algorithm}[h]
\caption{Raw Histogram Generation}\label{alg:hist}
\begin{algorithmic}
\Require 3D point cloud $\mathcal{P}$ and 2D accumulated event map $\mathcal{E}$
\Ensure 1D lidar hist, 1D event hist, 2D joint hist
\State Initialize: $\mathrm{hist_L}$, $\mathrm{hist_E}$, and $\mathrm{hist_{LE}}$
\For{$\left(\mathbf{P}_i, \mathrm{L}_i\right) \in \mathcal{P}$} \Comment 3D point \& intensity
\State $\boldsymbol{p}_i \gets  \mathbf{K}[\mathbf{R}|\boldsymbol{t}]\mathbf{P}_i$ \Comment 2D projection
\If{$\boldsymbol{p}_i$ in image bounds}
    \State $\mathrm{E}_i \gets \mathcal{E}(\boldsymbol{p}_i)$
    \State $\mathrm{hist_L}(\mathrm{L}_i) \gets \mathrm{hist_L}(\mathrm{L}_i) + 1$
    \State $\mathrm{hist_E}(\mathrm{E}_i) \gets \mathrm{hist_E}(\mathrm{E}_i) + 1$
    \State $\mathrm{hist_{LE}}(\mathrm{L}_i, \mathrm{E}_i) \gets \mathrm{hist_{LE}}(\mathrm{L}_i, \mathrm{E}_i) + 1$
\EndIf
\EndFor
\end{algorithmic}
\end{algorithm}

\begin{figure}
  \centering
  \includegraphics[width=0.8\columnwidth]{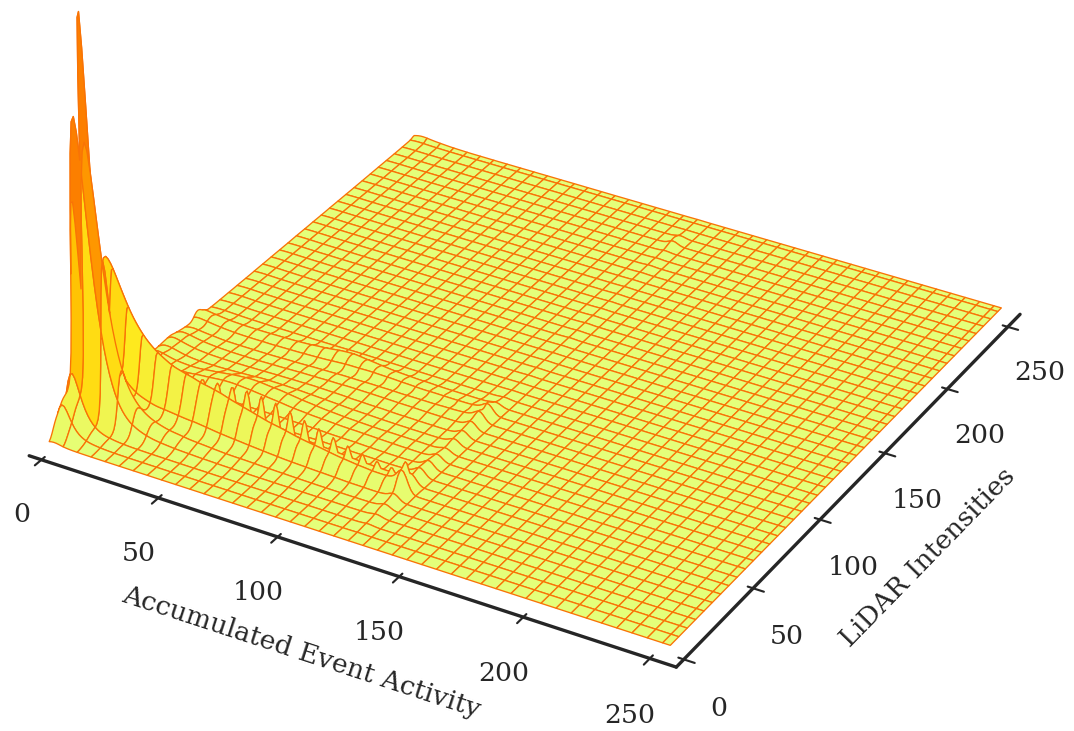}
  \caption{Smoothed joint histogram as the approximate joint probability distribution between the accumulated event map and lidar intensities.} \label{fig:hist2d}
\end{figure}

Subsequently, these raw histograms are normalized to acquire noisy approximations of the probability distribution functions $\hat{\mathrm{p}}_\mathrm{L}$, $\hat{\mathrm{p}}_\mathrm{E}$, and $\hat{\mathrm{p}}_\mathrm{LE}$. The histograms are normalized by $n$, the total number of points that project into the valid image region.
To suppress noise, we perform a kernel density estimation (KDE) using Silverman's rule-of-thumb~\cite{silverman1986density}. In practice, we apply Gaussian blurring convolutions, a method that is nearly equivalent to kernel density estimation (KDE) in the case where histogram bins are spaced equivalently to the discretized point values. 
Fig.~\ref{fig:hist2d} shows the joint probability distribution of the lidar intensities and the accumulated event map values.
The smoothed normalized histograms are finally used to get an MI estimate via Eq.~(\ref{eqn:mi})-(\ref{eqn:hxy}).

\subsection{Optimization Formulation}

With the formulated MI objective function and our approximations of the probability distributions, we formulate an optimization problem as follows:
\begin{equation}
\label{eqn:opt}
    \hat{\Theta} = \arg \max_\Theta \: \mathrm{MI}(\mathrm{L},\mathrm{E};\Theta)
\end{equation}
where $\Theta = (x, y, z, v_1, v_2, v_3)$ and MI is maximized at the correct extrinsic parameters given an arbitrary set of scenes. Effective optimization of this function is aided by greater smoothness and convexity in the objective function. 


\begin{figure*}[t]
  \centering
  \includegraphics[width=0.32\textwidth]{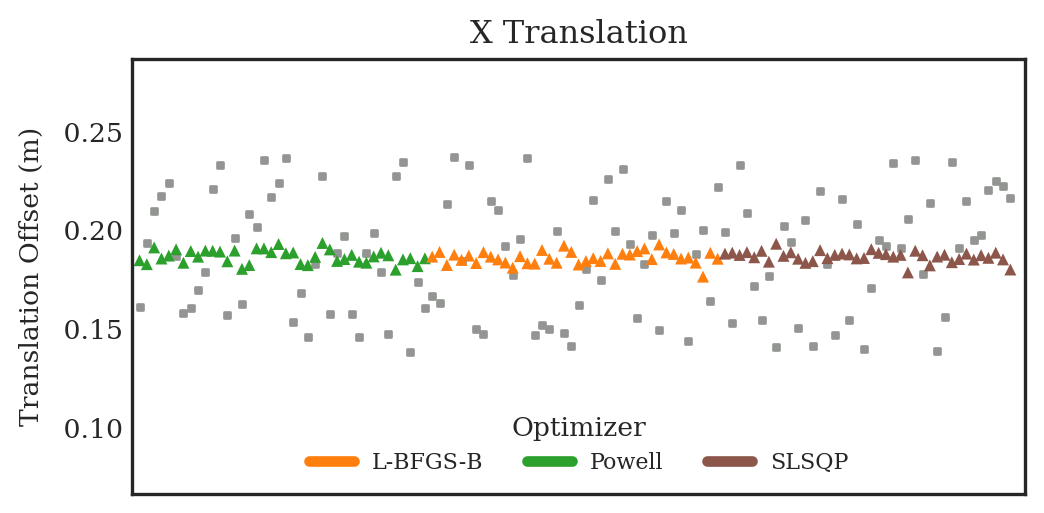}
  \hfill
  \includegraphics[width=0.32\textwidth]{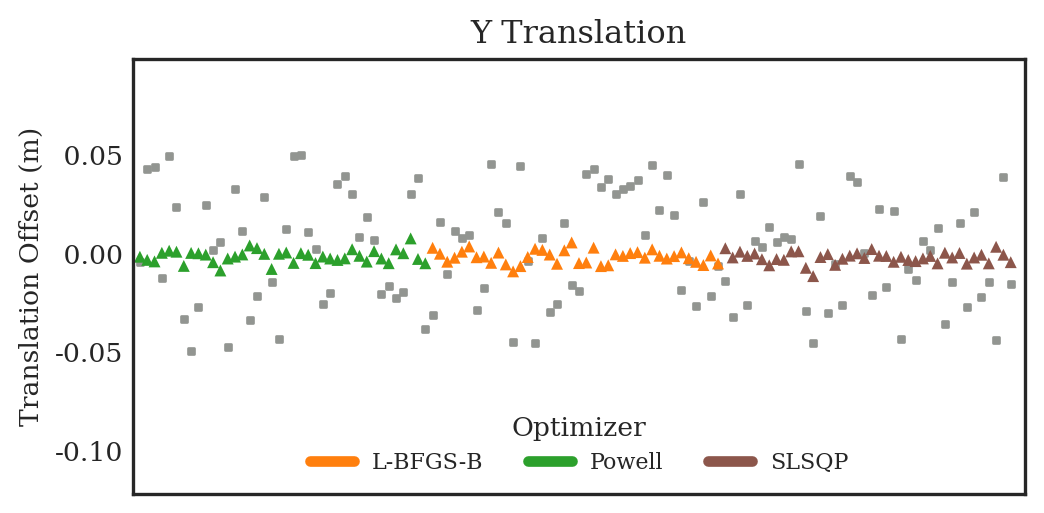}
  \hfill
  \includegraphics[width=0.32\textwidth]{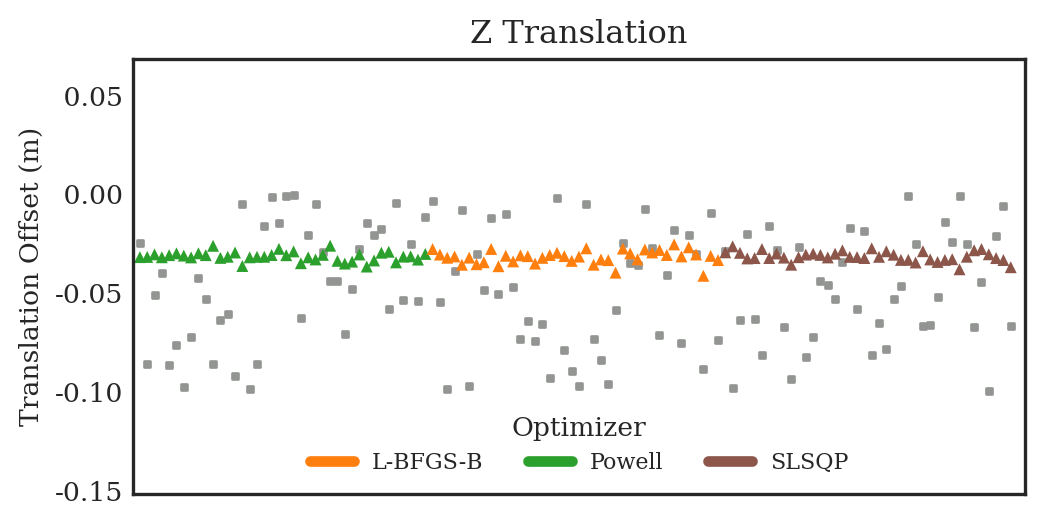}
  \includegraphics[width=0.32\textwidth]{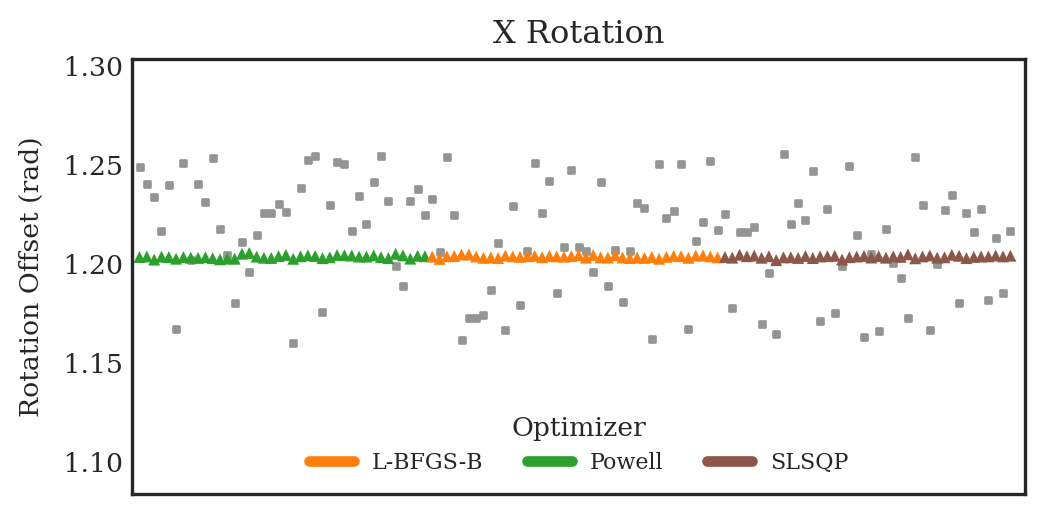}
  \hfill
  \includegraphics[width=0.32\textwidth]{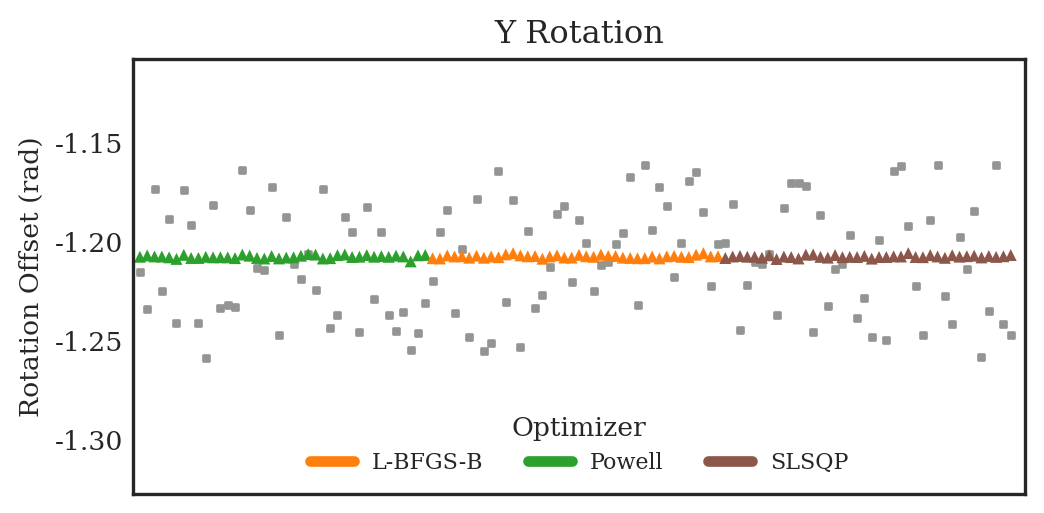}
  \hfill
  \includegraphics[width=0.32\textwidth]{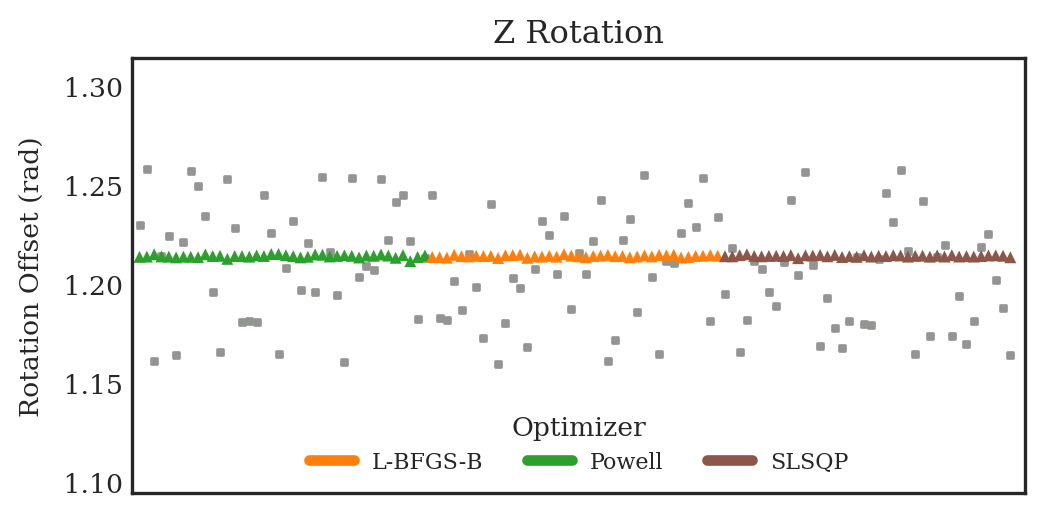}
  \caption{Optimized extrinsic calibration parameters for the translation and rotation where grey squares are the initial seed values. Best viewed on a screen.} \label{fig:aa-stable}
\end{figure*}

The optimization process can be theoretically accomplished by any numeric minimization algorithm. In this work, we show that multiple optimizers are effective. In particular we find that bounded numeric gradient-based methods are the most effective. Bounding is particularly critical to prevent the ill-posed optimization space where few or no points are projected into the accumulated event map. We implement these processes using the SciPy library~\cite{Virtanen2020SciPyPython}. 

\section{Experiments}

\subsection{Extrinsic Calibration Dataset}

Our calibration dataset consists of static scenes recorded over 3 seconds divided into two classes: checkerboard scenes and checkerless scenes. All scenes were collected in an underground garage. The checkerboard scenes were taken in a set position with a checkerboard target placed in close proximity to both the event camera and lidar. The checkerless scenes are captured in multiple locations without a checkerboard.
Our dataset consists of 35 checkerboard scenes and 58 checkerless scenes for a total of 93 scenes.

\subsection{Convergence Assessment}

We perform noise-induced experiments to evaluate robustness and run-time of the calibration for different optimizers. Specifically, we calculate the mean and standard deviation results of 40 calibrations, where the measured parameters are induced with uniform noise in the range of 0.1\,\si{m} and 0.1\,radians. Unless otherwise specified, 40 scenes are respectively sub-sampled from the total of 93 scenes.

\textbf{Robustness Analysis.}
Some tested optimizers failed to robustly converge. Specifically, the Nelder\hydash Mead simplex refinement method performed poorly in converging to the consensus translation parameters. Unbounded methods also occasionally converged in the ill-posed portion of the optimization space. 
We find that bounded Broyden\hydash Fletcher\hydash Goldfarb\hydash Shanno (L-BFGS-B), bounded Sequential Quadratic Programming (SLSQP), and the gradient-free Powell's Method perform well. Fig.~\ref{fig:aa-stable} shows the convergence robustness to seed errors for these optimizers. 
The convergence of calibration results in the presence of significant noise indicates a convergence basin robust against initialization errors. The rotation calibration is highly consistent, converging to results within a standard deviation of 0.0007\,rad (0.04°) against uniformly induced noise of 0.1\,rad (5.7°) for each axis. The translation calibration is also consistent, but exhibits a measurable standard deviation of 3\,\si{mm} against the 100\,\si{mm} of uniform noise induced in each translation axes. 

\begin{figure}
  \centering
  \includegraphics[width=0.9\columnwidth]{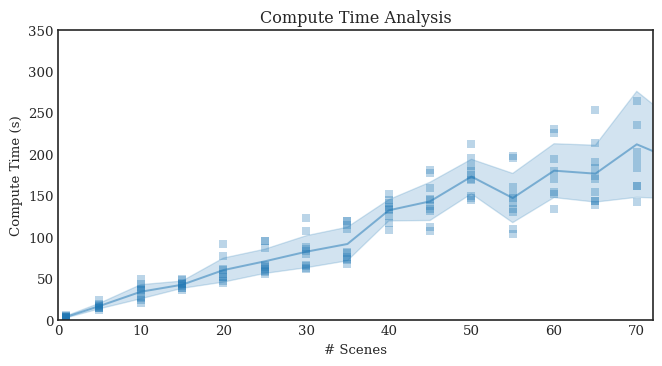}
  \caption{Computation time as a function of the number of scenes used in SLSQP optimization.} \label{fig:compute-scenes}
\end{figure}

\textbf{Computational Time Analysis.}
We find that of the three optimizers, SLSQP performs significantly faster than L-BFGS-B and Powell's Method. Powell's Method takes up to an order of magnitude longer than SLSQP and multiple times longer than \text{L-BFGS-B}. We thus choose SLSQP as the default optimizer for L2E. A full summary is shown in Table~\ref{tab:speed}.
\begin{table}
\caption{Average and standard deviation of run-times across 40 runs.} \label{tab:speed}
\centering
\renewcommand{\arraystretch}{1.3}
\begin{tabular*}{\columnwidth}{@{\extracolsep{\fill}}lccc@{}}
\toprule
\textbf{Optimizer} & Powell's Method & L-BFGS-B & SLSQP \\
\midrule
\textbf{Time\,(\si{s})}     & 934.4 $\pm$ 251.2 & 363.0 $\pm$ 94.2 & 134.2 $\pm$ 25.2 \\
\bottomrule
\end{tabular*}
\end{table}
We can also analyze the computational time in terms of the number of optimized scenes, $m$. More scenes lead to increased robustness, but longer computation. We can theorize that the optimization process has a complexity of $O(m)$, as the expected number of optimization steps should not depend on the scene count. This assumes that the underlying cost landscape structure is independent of the number of scenes used as scene sampling is a random process. Consequently, the computation depends on the mutual information calculation, which must calculate the projection and mutual information for each individual scene. This calculation should scale linearly with the number of scenes used, which is observed in our experiments shown in Fig.~\ref{fig:compute-scenes}.

\begin{figure}
  \centering
  \includegraphics[width=0.49\columnwidth]{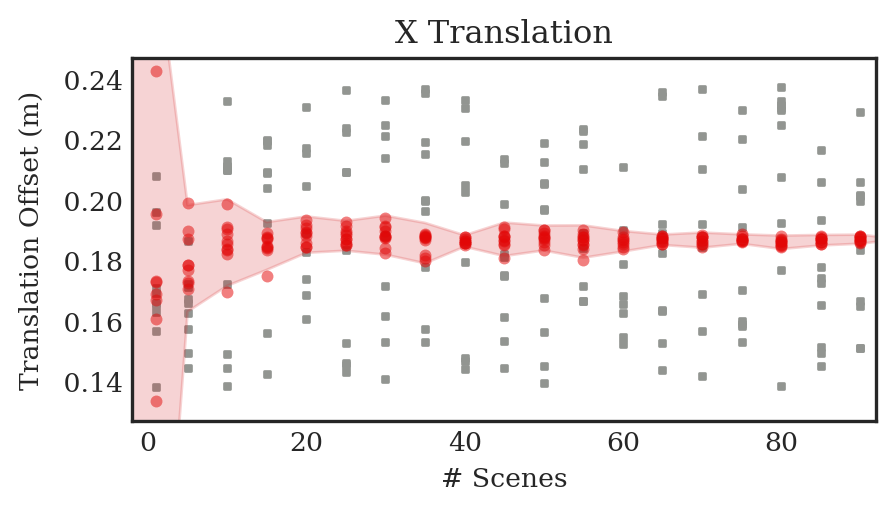}
  \hfill
  \includegraphics[width=0.49\columnwidth]{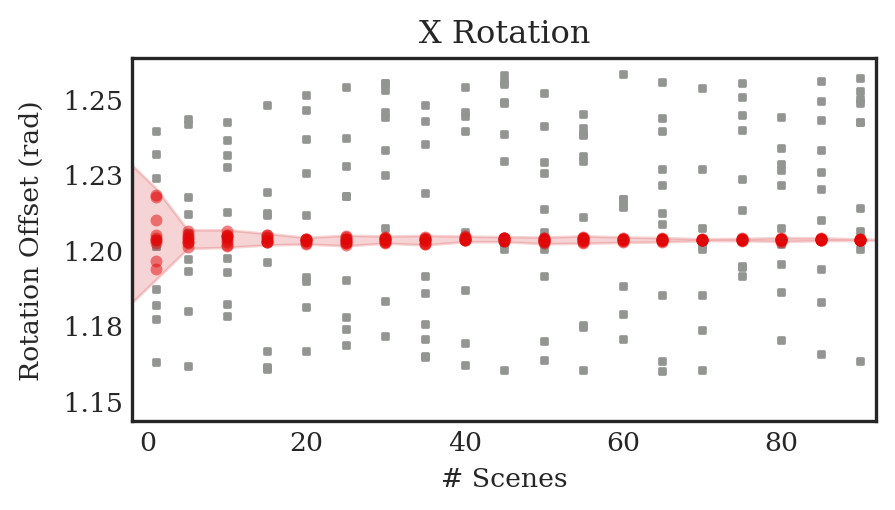}
  \includegraphics[width=0.49\columnwidth]{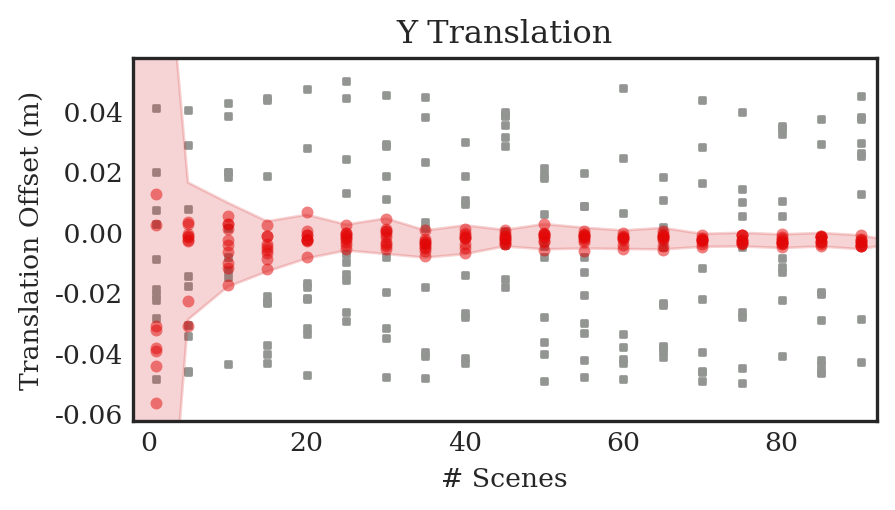}
  \hfill
  \includegraphics[width=0.49\columnwidth]{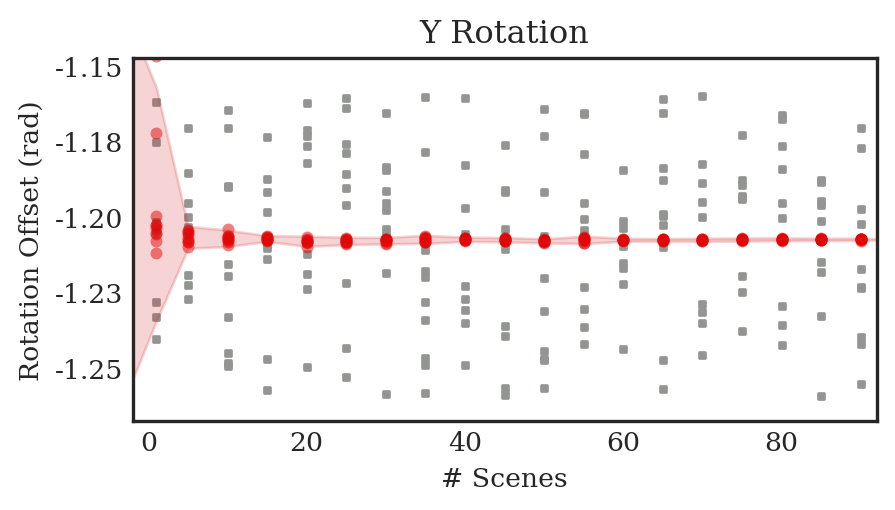}
  \includegraphics[width=0.49\columnwidth]{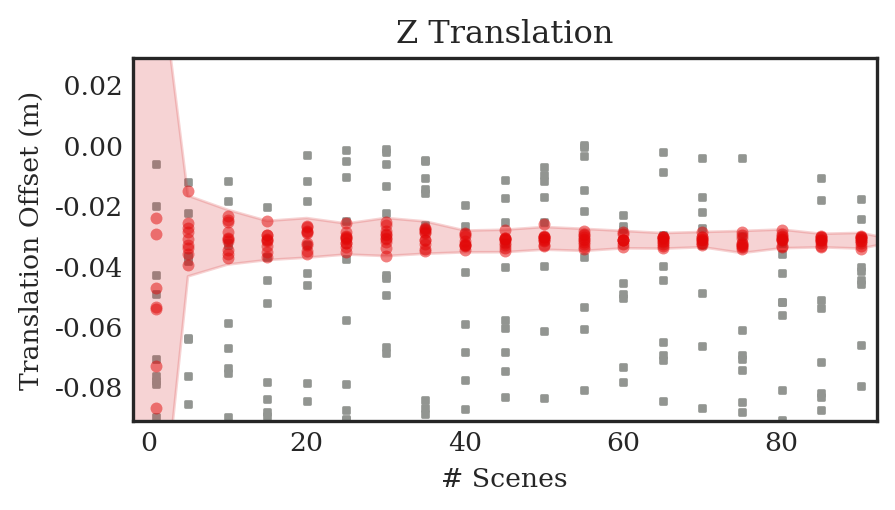}
  \hfill
  \includegraphics[width=0.49\columnwidth]{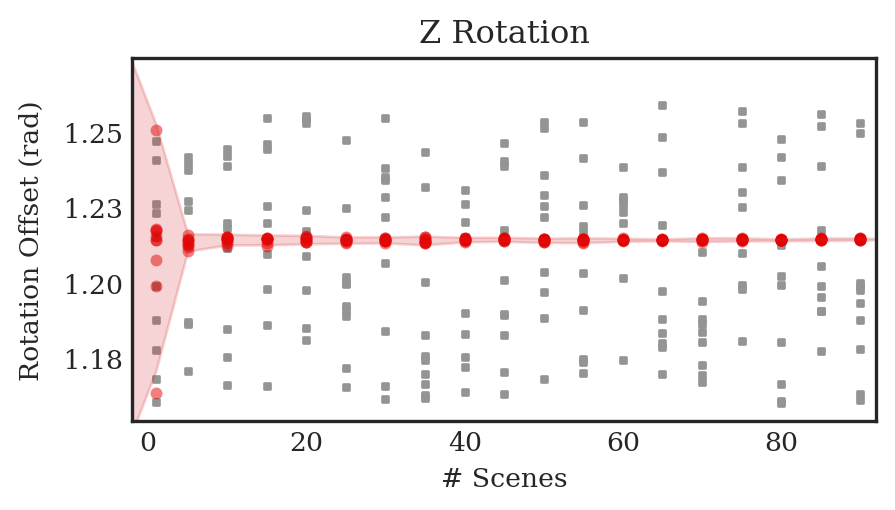}
  \caption{Calibration results in red from SLSQP optimization for increasing number of scenes from left to right. Grey squares represent the initial seeds.} \label{fig:scene-num}
\end{figure}

\textbf{Impact of Number of Scenes.} One of the key components for optimization is the collection of sufficient scenes for optimization. We investigate the number of scenes required for consistent calibration in Fig.~\ref{fig:scene-num}. As expected, increasing the number of scenes used in optimization decreases repeatability variance. The variance in calibration has substantially decreased when using 20 or more scenes in the optimization. 

\subsection{Performance Comparison}

\textbf{Checkerboard \textit{vs.}\ Checkerless Scenes.} We perform repeated experiments using 20 sub-sampled scenes within the full, checkerboard-only, and checkerless subsets. We evaluate the per-scene MI for held-out scenes. Results from these calibrations are shown in Table~\ref{tab:mi-values}, rows 1-3. We report marginally higher average MI on the full evaluation when using the checkerboard-only subset \vs checkerless-only subset. 

\begin{table}
\caption{Averaged MI scores comparison between our calibration procedure (MI optimization), the two-stage baseline, and the uncalibrated results.} \label{tab:mi-values}
\centering
\renewcommand{\arraystretch}{1.3}
\begin{tabular*}{\columnwidth}{@{\extracolsep{\fill}}lrrr@{}}
\toprule
Calibration & Full Eval. & Checker Eval. & Checkerless Eval. \\
\midrule
Ours (Mixed Set)    & \textbf{0.44119} & 0.53282 & 0.36729 \\
Ours (Checker-only) & 0.44082 & \textbf{0.53421} & 0.36550 \\
Ours (Checkerless)  & 0.44064 &  0.53092 & \textbf{0.36782} \\
2-Stage Baseline     & 0.41794 &  0.51664 &  0.33834 \\
Uncalibrated & 0.28767 &  0.33834 &  0.23873 \\
\bottomrule
\end{tabular*}
\end{table}

\begin{figure}
  \centering
  \setlength{\tabcolsep}{2pt}
  \begin{tabularx}{\textwidth}{lr}
    \includegraphics[width=0.49\columnwidth]{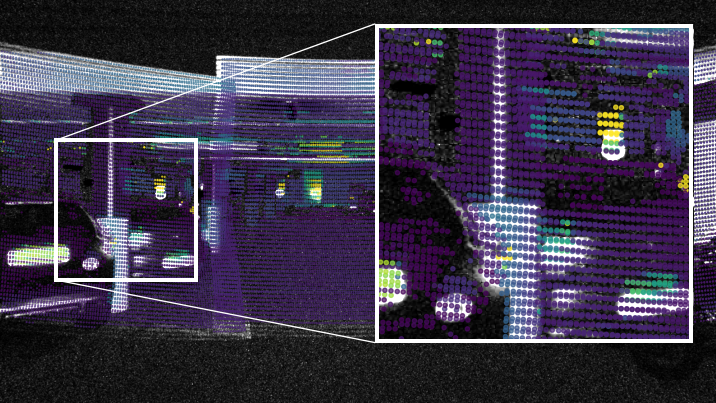} &
    \includegraphics[width=0.49\columnwidth]{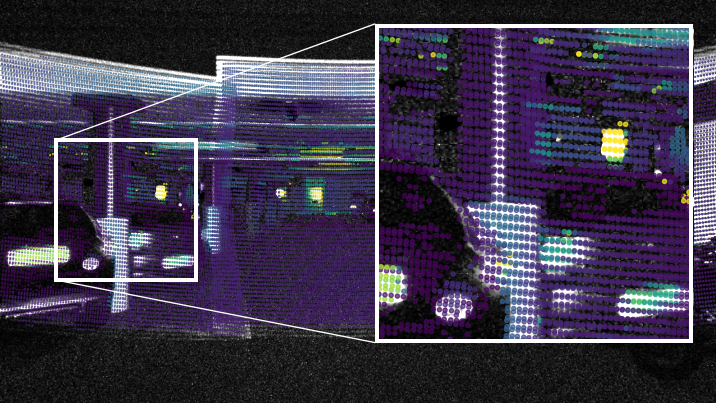} \\
    \includegraphics[width=0.49\columnwidth]{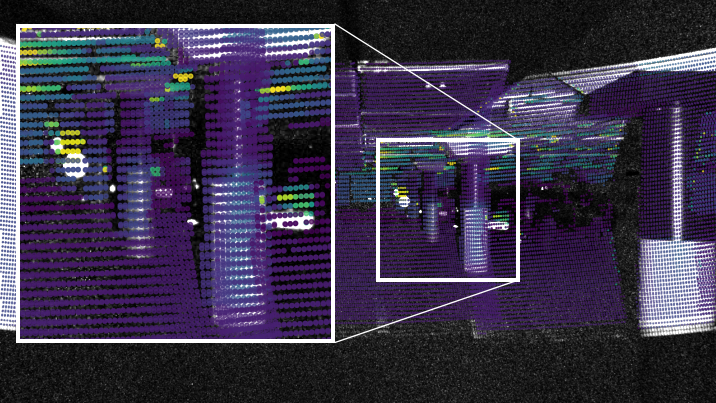} &
    \includegraphics[width=0.49\columnwidth]{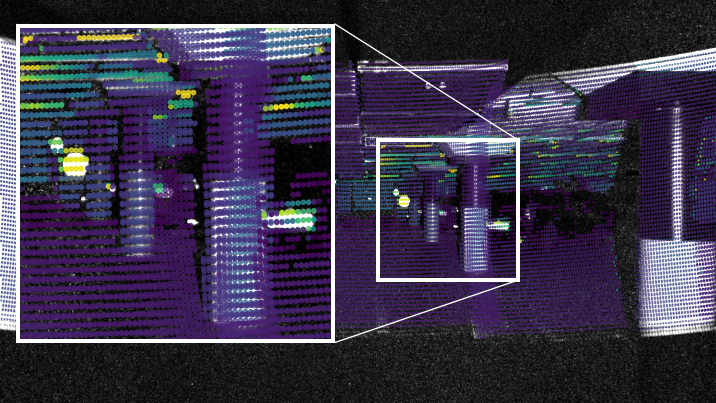} \\
    \includegraphics[width=0.49\columnwidth]{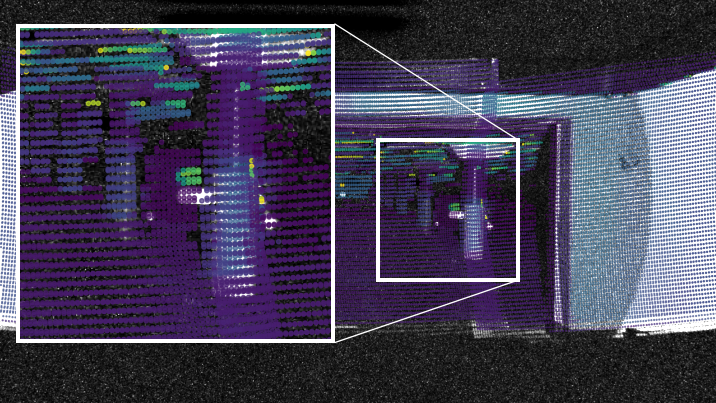} &
    \includegraphics[width=0.49\columnwidth]{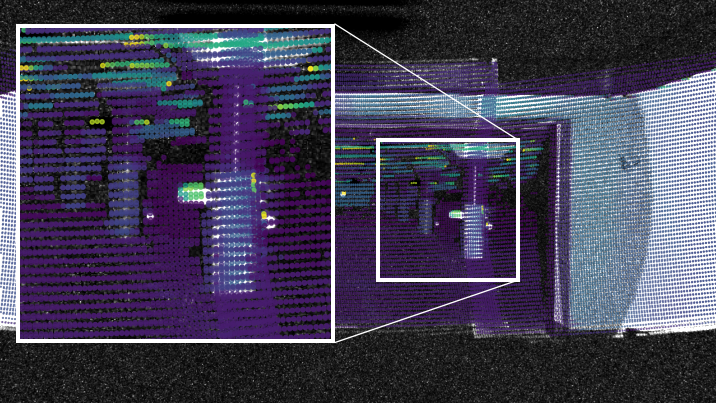}
  \end{tabularx}
  \caption{(Left) Baseline. (Right) L2E calibration. Best viewed on a screen.} \label{fig:2calibrated}
\end{figure}

\textbf{Two-stage Calibration Comparison.} We compare our method to a two-stage method that first stereo-calibrates the event camera to a standard camera intermediary requiring time-synchronized and motion-dependent event-to-video reconstruction \cite{Muglikar2021} and then calibrates to the lidar using the Lidar-Camera Toolbox~\cite{Zhou2018AutomaticCorrespondences}. Our method achieves better visual alignment, as evidenced by Fig.~\ref{fig:2calibrated}, and better information agreement, as seen in Table~\ref{tab:mi-values}, row 4. The two stage-process propagates errors for each calibration process (lidar to camera, camera to event camera), leading to greater errors in the indirectly connected sensors.

\section{Discussion}

\textbf{Overcoming time-synchronization and motion dependence.} A key benefit of our approach is overcoming challenges that make traditional methods of calibration impossible. Event-driven sensors are incapable of providing information in static scenes, thus demanding temporal synchronization and scene motion for extrinsic calibration. In contrast, our approach decouples the effects of time synchronization from geometric calibration.

\textbf{Considerations in Scene Selection.} L2E can be used in an unstructured environment with any combination of static scenes. However, we find that L2E tends to perform slightly better when  observing high-texture objects\textemdash \ie our checkerboards. This phenomenon is likely a result of the proximity of the checkerboard targets to both sensors, providing stronger rotation and translation constraints to the optimization. Poorer optimization in outdoor scenes was observed in~\cite{Pandey2014}, where errors were partially attributed to having fewer near-field 3D points in outdoor scenes. Consequently, the use of close-proximity objects with a reasonable amount of texture should have a beneficial effect on the calibration results. We note that precise geometric controls\textemdash \eg checkerboard target tolerances\textemdash are not required, as opposed to traditional camera calibration methods. 


\textbf{Managing NIR Sensitivity.} Our method relies on the registration of lidar laser signals by the event camera. However, NIR sensitivity may be undesirable during standard operation. To address this concern, the Prophesee GEN4.1 contains bias tuning parameters that can eliminate high frequency flicker effects, which we have found to be able to eliminate registration of lidar pulses. 

\section{Conclusion}

This work presents L2E, the first direct extrinsic calibration method between event-based cameras and lidars using event-based structured light. L2E offers a flexible automatic approach to extrinsic calibration that leverages direct correlation between the lidar active signals and the corresponding registered events without the need for precise time synchronization or dynamic scenes. As event-driven vision technology continues to mature, we hope that our direct calibration method will enable further research on event cameras and sensor fusion.

\clearpage
{
\bibliographystyle{IEEEtran}
\bibliography{references}
}

\end{document}